\documentclass[sigconf]{acmart}

\usepackage{graphicx}
\usepackage{subcaption}
\usepackage{wrapfig}
\usepackage{multirow}
\usepackage{makecell}
\usepackage{amsmath}
\usepackage{listings}
\usepackage{xspace}
\usepackage{enumitem} 
\newtheorem{definition}{Definition}
\usepackage{xcolor}              
\usepackage{amsmath,amsfonts}
\usepackage{algorithmic}
\usepackage{graphicx}
\usepackage{textcomp}
\usepackage{xcolor}
\usepackage{subcaption}
\usepackage[linesnumbered,ruled,vlined]{algorithm2e}

\definecolor{RoyalBlue}{RGB}{0,45,200}
\definecolor{InputColor}{RGB}{0,70,0}

\newcolumntype{L}[1]{>{\raggedright\let\newline\\\arraybackslash}p{#1}}
\newcommand{\psar}{pSAR\xspace}

\newcommand{\MyFrameBox}[1]{
  \par\addvspace{4pt}
  \begingroup
  \setlength{\fboxrule}{0.2pt}
  \setlength{\fboxsep}{2pt}
  \noindent\fcolorbox{gray!70}{gray!15}{
    \begin{minipage}{\linewidth}
      \small 
      #1
    \end{minipage}
  }
  \endgroup
  \par\addvspace{2pt}
}

\acmYear{2026}
\acmConference[CAIN]{CAIN 2026}{April 2026}{Rio de Janeiro, Brazil}
\acmBooktitle{Proceedings of CAIN 2026}

\copyrightyear{2026}
\acmYear{2026}
\setcopyright{cc}
\setcctype{by}
\acmConference[CAIN '26]{2026 IEEE/ACM 5th International Conference on AI Engineering - Software Engineering for AI}{April 12--13, 2026}{Rio de Janeiro, Brazil}
\acmBooktitle{2026 IEEE/ACM 5th International Conference on AI Engineering - Software Engineering for AI (CAIN '26), April 12--13, 2026, Rio de Janeiro, Brazil}
\acmPrice{}
\acmDOI{10.1145/3793653.3793768}
\acmISBN{979-8-4007-2475-6/2026/04}

\begin{document}

\title[Cognition Envelopes for Bounded Decision Making in sUAS]{Cognition Envelopes for Bounded Decision Making \\in Autonomous UAS Operations}

\author{Pedro Antonio Alarc\'on Granadeno}
\email{palarcon@nd.edu}
\orcid{0009-0006-7829-7088}
\affiliation{%
  \institution{University of Notre Dame}
  \department{Computer Science and Engineering}
  \city{Notre Dame}
  \state{IN}
  \country{USA}
}

\author{Arturo Miguel Russell Bernal}
\email{arussel8@nd.edu}
\orcid{0009-0009-2902-5766}
\affiliation{%
  \institution{University of Notre Dame}
  \department{Computer Science and Engineering}
  \city{Notre Dame}
  \state{IN}
  \country{USA}
}

\author{Sofia Nelson}
\email{snelso24@nd.edu}
\orcid{0009-0004-5835-5047}
\affiliation{%
  \institution{University of Notre Dame}
  \department{Computer Science and Engineering}
  \city{Notre Dame}
  \state{IN}
  \country{USA}
}

\author{Demetrius Hernandez}
\email{dhernan7@nd.edu}
\orcid{0000-0002-7813-8478}
\affiliation{%
  \institution{University of Notre Dame}
  \department{Computer Science and Engineering}
  \city{Notre Dame}
  \state{IN}
  \country{USA}
}

\author{Maureen Petterson}
\email{mpetters@nd.edu}
\orcid{0009-0008-8801-0751}
\affiliation{%
  \institution{University of Notre Dame}
  \department{Computer Science and Engineering}
  \city{Notre Dame}
  \state{IN}
  \country{USA}
}

\author{Michael Murphy}
\email{mmurph51@nd.edu}
\orcid{0009-0007-4303-412X}
\affiliation{%
  \institution{University of Notre Dame}
  \department{Computer Science and Engineering}
  \city{Notre Dame}
  \state{IN}
  \country{USA}
}

\author{Walter J. Scheirer}
\email{walter.scheirer@nd.edu}
\orcid{0000-0001-9649-8074}
\affiliation{%
  \institution{University of Notre Dame}
  \department{Computer Science and Engineering}
  \city{Notre Dame}
  \state{IN}
  \country{USA}
}

\author{Jane Cleland-Huang}
\email{janehuang@nd.edu}
\orcid{0000-0001-9436-5606}
\affiliation{%
  \institution{University of Notre Dame}
  \department{Computer Science and Engineering}
  \city{Notre Dame}
  \state{IN}
  \country{USA}
}

\begin{abstract}
Cyber-physical systems increasingly rely on foundational models, such as Large Language Models (LLMs) and Vision-Language Models (VLMs) to increase autonomy through enhanced perception, inference, and planning. However, these models also introduce new types of errors, such as hallucinations, over-generalizations, and context misalignments, resulting in incorrect and flawed decisions. To address this, we introduce the concept of Cognition Envelopes, designed to establish reasoning boundaries that constrain AI-generated decisions while complementing the use of meta-cognition and traditional safety envelopes. As with safety envelopes, Cognition Envelopes require practical guidelines and systematic processes for their definition, validation, and assurance. In this paper we describe an LLM/VLM-supported pipeline for dynamic clue analysis within the domain of small autonomous Uncrewed Aerial Systems deployed on Search and Rescue (SAR) missions, and a Cognition Envelope based on probabilistic reasoning and resource analysis. We evaluate the approach through assessing decisions made by our Clue Analysis Pipeline in a series of SAR missions. Finally, we identify key software engineering challenges for systematically designing, implementing, and validating Cognition Envelopes for AI-supported decisions in cyber-physical systems.
\end{abstract}


\maketitle
\renewcommand{\shortauthors}{Alarc\'on Granadeno, Russell Bernal, Nelson, Hernandez, Petterson, Murphy, Scheirer, Cleland-Huang}
% Sections
\section{Motivation}
Advances in Large Language Models (LLMs) and Vision-Language Models (VLMs) have laid the foundations for improved perception, inference, and planning for autonomous Cyber-Physical Agents, such as small Uncrewed Aerial Systems (sUAS) \cite{llm4uavsurvey, hu2025llvm,seams-keynote2025}. At the same time, the growing autonomy enabled by these models heightens the risk that AI errors, such as hallucinations, result in incorrect understanding and flawed decision-making~\cite{openai2025hallucinate, wang2024hallucination}. In domains such as aerial Search and Rescue (SAR), where autonomous sUAS are deployed in life-critical operations, such failures can compromise mission objectives, endanger human lives, and erode trust \cite{DBLP:conf/chi/AgrawalABCFHHTK20,Moore2024IASTesting}. To address these challenges, we introduce \emph{Cognition Envelopes}, designed to detect flawed decisions that occur when reasoning outcomes contradict available evidence, violate operational constraints, or lack evidential support due to model errors or hallucinations.

\begin{figure}[t]
    \centering
    \includegraphics[width=\linewidth]{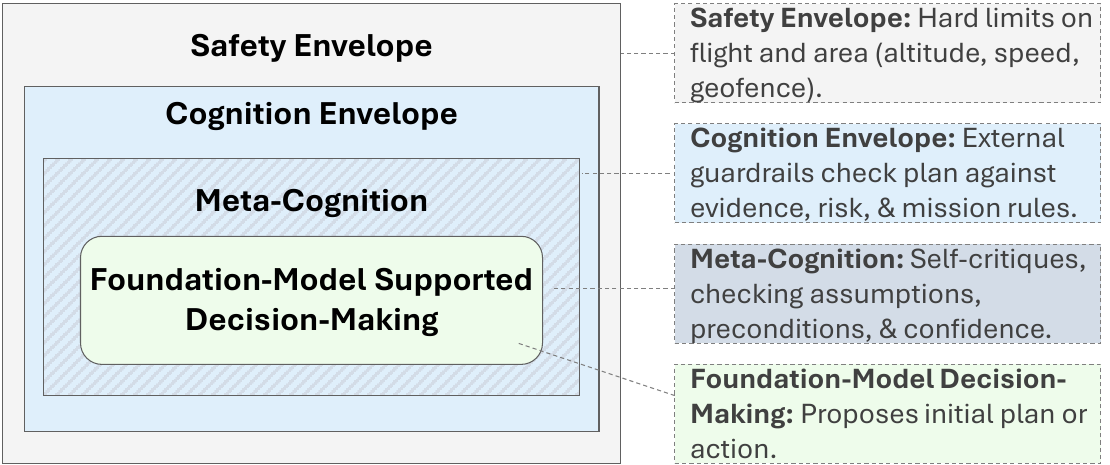}
    \caption{Layered decision-making guardrails: the outer Safety Envelope sets hard limits, while the Cognition Envelope applies external guardrails on inner-level decisions.}
    \label{fig:cognition-hierarchy}
\end{figure}

As illustrated in Figure \ref{fig:cognition-hierarchy}, Cognition Envelopes are related to, but distinct from, several established mechanisms. For instance, meta-cognition operates within the model's own reasoning loop when a model self-critiques and refines its own plans \cite{conway2024toward, shinn2023reflexion}. While useful, meta-cognition remains coupled to the same generative process that produced the original decision and can therefore inherit the same blind spots or hallucinated premises \cite{lu2025auditingmetacognitivehallucinationsreasoning}. In contrast, Cognition Envelopes explicitly require external, independent monitors to validate a decision. Safety envelopes in cyber-physical autonomy typically bound physical state and control within certified limits (e.g., geofences, separation constraints, or safe operating regions) and primarily prevent unsafe operations \cite{10.1007/978-3-319-74183-3_1}. Cognition Envelopes instead operate at the decision semantics level, where a plan can remain physically admissible and safe, yet still be wrong for the mission because it contradicts evidence, misallocates scarce resources, or pursues an unjustified objective.

While safety envelopes are supported by a mature ecosystem of standards and assurance frameworks~\cite{IEC61508,ISO26262,DO178C,ARP4754A,UL4600,LevesonThomas2018,ISO21448,Schaefer2022,Leveson2011,Rushby2019}, no comparable guidance exists for defining, validating, or assuring \emph{Cognition Envelopes}. This paper therefore introduces the concept of \emph{Cognition Envelopes}, illustrates their practical use for AI-supported reasoning in the domain of Search and Rescue (SAR) with small Uncrewed Aerial Systems (sUAS), and lays out preliminary guidance and open software engineering challenges related to integrating them in software intensive systems.  We formalize this concept as follows:

\begin{definition}[Cognition Envelope]
A Cognition Envelope is a runtime assurance layer that gates execution of a foundation-model pipeline using \textbf{semantic acceptability criteria} grounded in \textbf{external evidence}, uncertainty, and resource risk. Formally, we describe a Cognition Envelope as a tuple $\mathcal{E}=\langle d, e, M, s, G\rangle$, where $d$ is the candidate decision produced by the foundation-model pipeline, $e$ is the \textbf{external evidence} and runtime context available at decision time, $M$ is an \textbf{external semantic model} that evaluates $d$ under $e$ to produce a semantic acceptability signal $s:=M(d,e)$ (e.g., a score, a constraint-satisfaction outcome, or a set of triggered checks), and $G$ is a gating rule that uses $s$ to determine whether the system may enact $d$ (or must defer, revise, or escalate).
\end{definition}

The discussion of Cognition Envelopes is timely given the growing body of work in the robotics space, where researchers have proposed the use of LLM and VLM-based techniques for diverse tasks such as image interpretation, intelligent task assignment, and vision-based trajectory planning \cite{VLM-Social-Nav, VLM-GroNav, LLM-Drone-Nav, khan2025context, VLM-Drone-Indoor}. However, this research area of AI autonomy has progressed with little to no focus on associated software engineering practices and challenges. 

To describe and explore the use of Cognition Envelopes, we start by presenting a concrete application within the sUAS SAR domain where a {\it Clue Analysis Pipeline} (CAP) is used to manage visual clues, such as discarded items or footprints detected by sUAS during a search. The CAP utilizes multi-modal foundational models to analyze the captured image of a clue, determine its relevance, and plan a subsequent action, $d$. We integrate two safeguards into our Cognition Envelope, which together serve as our \textit{external semantic validator}, $M$. The first is a Probability-Based SAR Model ({\it \psar}) that computes the probability of a lost person being found in any given area of the search region at a given stage of an unfolding mission. The second is a simple Mission Cost Evaluator (MCE) that examines the cost of executing a search plan. Together, the \psar and the MCE  are responsible for semantically assessing, and potentially constraining plans, generated by the CAP. Our worked example from the SAR domain not only illustrates the use of a Cognition Envelope in practice, but also serves as a vehicle for exploring the Software Engineering challenges involved in developing such systems, spanning requirements elicitation, architectural design, validation, and edge-based deployment under resource constraints. 

\begin{figure}[]
    \centering

\begin{subfigure}[t]{0.55\linewidth}
    \centering
    \includegraphics[width=\linewidth]{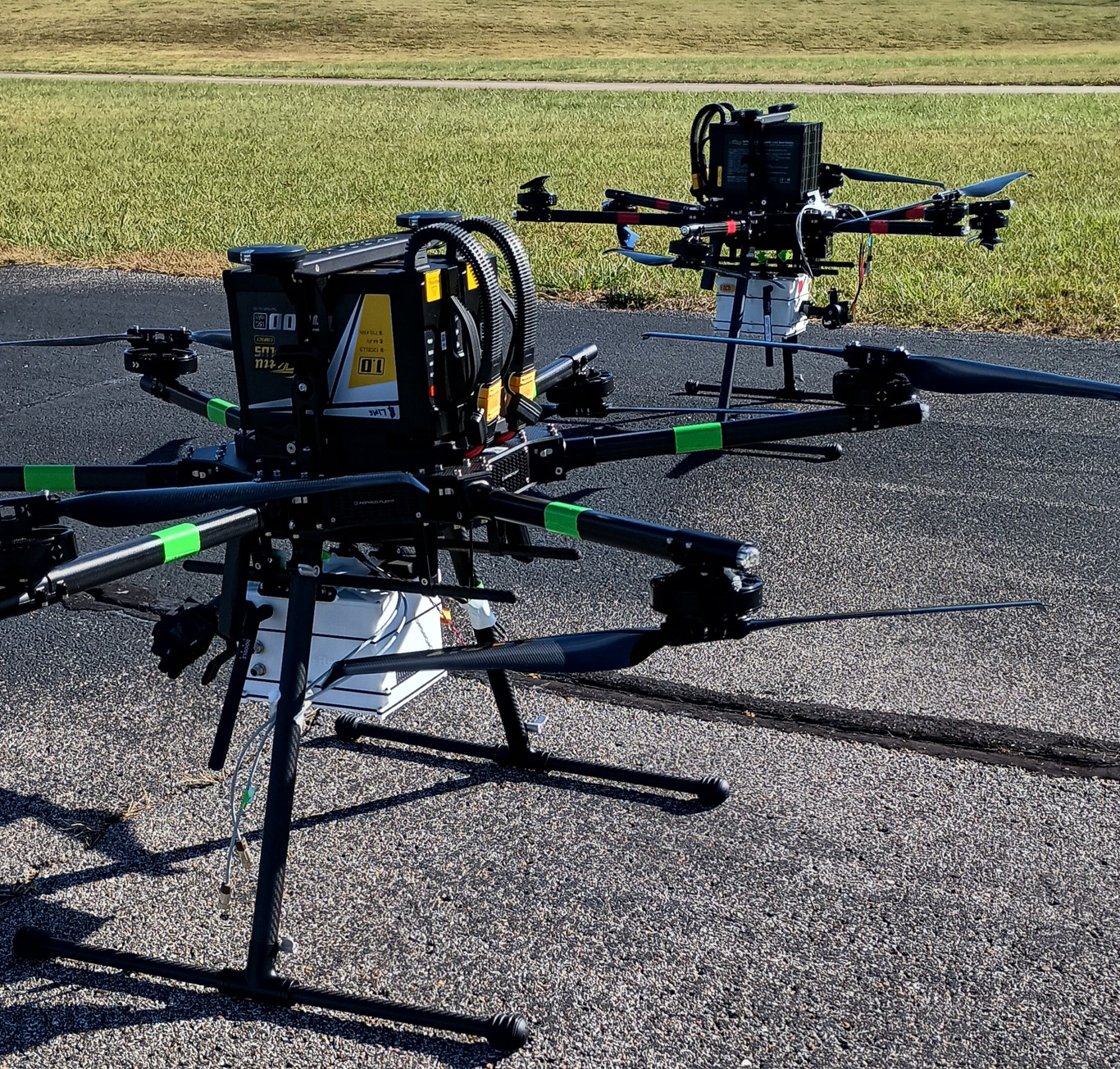}
    \subcaption{Two hexacopters are dispatched to search for a missing person.}
    \label{fig:drones}
\end{subfigure}
\hfill
    \begin{subfigure}[t]{0.402\linewidth}
    \centering
    \includegraphics[width=\linewidth,trim={20 0 30 0},clip]{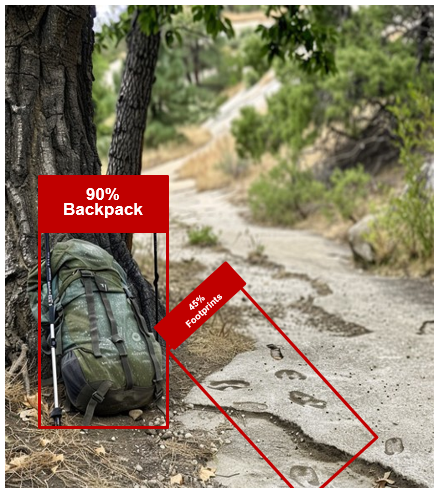}
    \subcaption{A backpack serves as a clue during a SAR mission.}
    \label{fig:backpack}
\end{subfigure}
\vspace{-8pt}
\caption{An sUAS discovers a backpack at the trailhead while conducting a trail-based search for a lost hiker. This clue could trigger mission-level adaptations.}
\label{fig:scenario}
\vspace{-12pt}
\end{figure}

%The remainder of this paper is organized as follows. Section~\ref{sec:clues} presents the LLM/VLM pipeline that supports clue analysis. Sections~\ref{sec:cognition_envelope} and~\ref{sec:validation} describe and evaluate the use of a Cognition Envelope, including an assessment of its generalizability. Section~\ref{sec:roadmap} outlines a set of open Software Engineering challenges, and Sections~\ref{sec:related}–\ref{sec:conclusions} discuss related work, threats to validity, and conclusions.

\section{Clue Analysis using Foundational Models}
\label{sec:clues}
In this paper we focus on one aspect of SAR, namely the detection of a clue by the sUAS, and its subsequent analysis using foundational AI models.   Figure \ref{fig:scenario} illustrates this scenario, showing two sUAS ready for dispatch on a mission, where they ultimately discover a discarded backpack, serving as a critical clue. The scenario assumes an ongoing mission in which sUAS are executing search tasks assigned by an automated mission planner (e.g., \cite{granadeno2025coveragepathplanningholonomic}).  In this scenario, each sUAS autonomously perceives its surroundings using onboard computer vision to detect candidate clues in open-world settings. In our implementation, we use the YOLO-World model for open-vocabulary detection tasks \cite{cheng2024yolo}. When a potential clue is detected, the system estimates its location \cite{bernal2025terrainvalidation} and extracts a representative image frame that is passed to the CAP.

To support both clue geolocation and task planning, the search region is represented as a model of the scene derived via multi-source data fusion. We combine satellite imagery with public geospatial datasets, including Digital Terrain Maps (DTMs) for elevation~\cite{thatcher2020usgs}, hydrographic features \cite{usgs_nhdplushr}, forest ecosystems \cite{usgs_annual_nlcd_2025}, and road/trail networks \cite{usgs_transportation_2023}. Satellite imagery is acquired via commercial APIs and processed by a convolutional neural network (CNN) trained to classify terrain categories. CNN and USGS labels are cross-referenced to resolve conflicts and fill gaps, deferring to USGS where reliable and to CNN where USGS data are missing or outdated. The resulting terrain is discretized into a uniform grid of cells $C$, where each cell $c \in C$ is assigned a dominant terrain label $\tau(c)$ (e.g., \emph{trail}, \emph{road}, \emph{lake}, \emph{river/stream}, \emph{forest}, etc.). Spatially contiguous cells with the same label form a terrain cluster $K_i$, and large clusters are further partitioned via $k$-means into compact search subclusters $\kappa \in \mathcal{K}$, which we treat as the named search areas for the mission.

In this paper, we plan at the level of terrain subclusters rather than individual cells. Subclusters are compact regions derived from the terrain grid that are large enough to serve as practical search areas. CAP therefore outputs one or more proposed subclusters $\kappa \in \mathcal{K}$ as its recommendation of \emph{where to search next}. We treat this recommendation as the candidate decision $d$ (Definition~1), which the Cognition Envelope evaluates against the available evidence and mission context before enactment. Throughout the remainder of the paper, we use the terms \emph{terrain subcluster}, \emph{named search area}, and \emph{search candidate} interchangeably to refer to $\kappa \in \mathcal{K}$, or equivalently decision $d$. 

\subsection{AI-Supported Clue Analysis Pipeline}
\begin{figure}
    \centering
    \includegraphics[width=\linewidth]{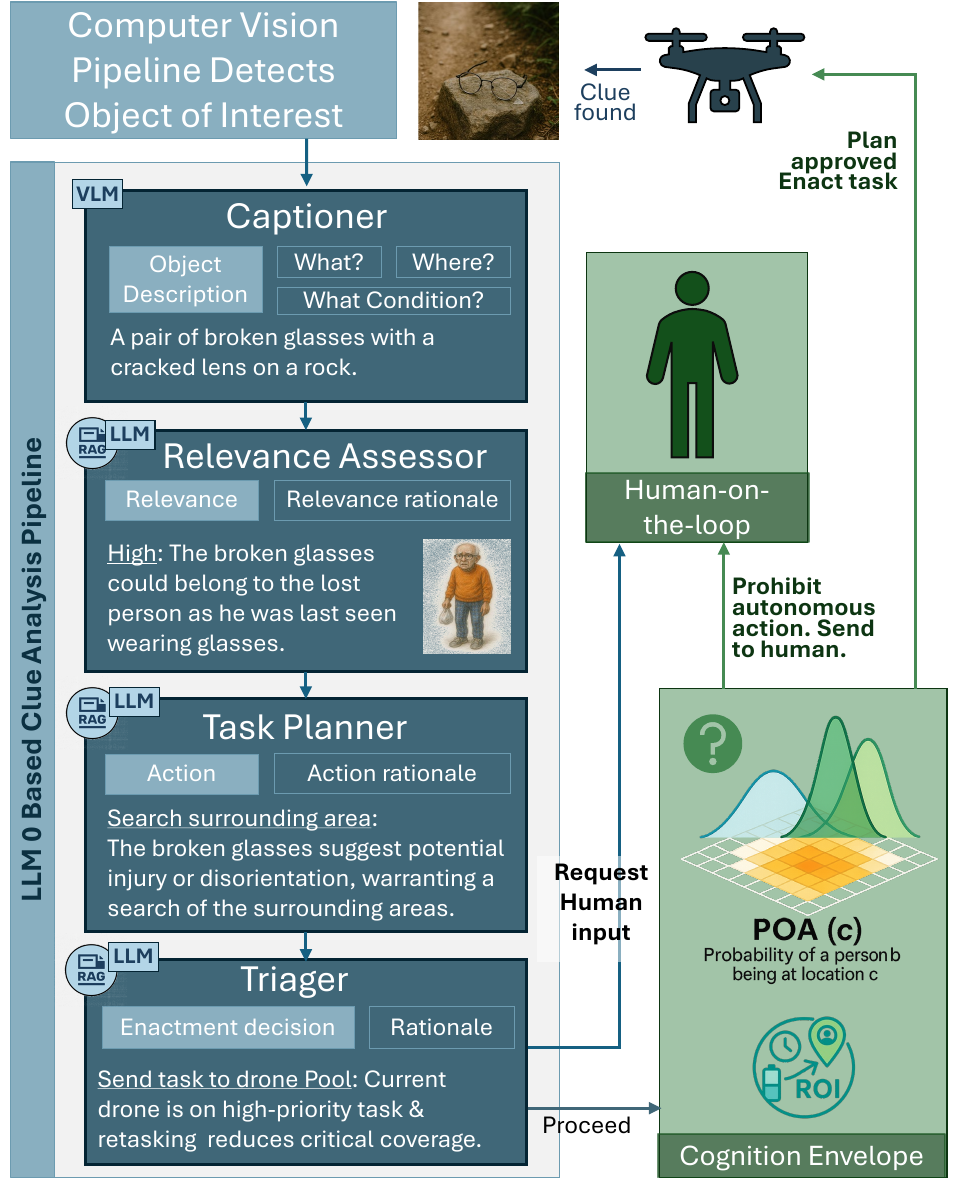}
    \caption{The Clue Analysis Pipeline includes four stages. The VLM generates a caption for the clue, while RAG + LLM analyzes its relevance. The Task Planner and Triager then decide whether to enact, revise, or defer the resulting action to a human operator. The Cognition Envelope approves the task or redirects it to a human for further assessment. }
    \label{fig:pipeline}
    %\vspace{-18pt}
\end{figure}

The CAP pipeline consists of four primary stages responsible for interpreting and captioning an image of a clue, assessing its relevance, planning appropriate tasks, and deciding whether a human should vet the generated plan. Each stage of the CAP pipeline is LLM-enabled using GPT-4o and includes a set of inputs, a carefully engineered prompt, and a set of outputs. In addition, all stages, except the first, use Retrieval Augmented Generation (RAG) to integrate external guidance into the prompt. For illustrative purposes we describe these stages using the example depicted in Figure \ref{fig:pipeline}, with additional information provided as supplemental material.  

\begin{itemize}[leftmargin=*, itemsep=2pt]
    \item{\bf Pipeline Trigger:}
The pipeline is triggered when the sUAS sends a geolocated image frame of a clue to the pipeline.
\item{\bf Stage 1 - Captioner:}  The Captioner receives the clue as a single image frame and is prompted to generate a structured description. The interaction involves the following inputs (green) and outputs (blue). Colored text is shown verbatim from text used within the prompt and from the generated responses.

\MyFrameBox{
\centering
\addtolength{\tabcolsep}{-3.8pt}
\begin{tabular}{@{}rL{6.6cm}@{}}
\textbf{Image:} & \textcolor{InputColor}{A single image frame}
\vspace{4pt} \\
\textbf{Clue:} & \textcolor{RoyalBlue}{A pair of broken glasses with a cracked lens on a rock.}\\
\textbf{What?} & \textcolor{RoyalBlue}{A pair of broken glasses}\\
\textbf{Where?} & \textcolor{RoyalBlue}{On a rock}\\
\textbf{Condition?} & \textcolor{RoyalBlue}{Cracked lens}
\end{tabular}
}

The captioner outputs a structured description of the clue for use in subsequent pipeline steps.

\item{\bf Stage 2 - Relevance Checker:} The relevance checker, as the name suggests, is responsible for assessing the relevance of the clue with respect to the lost person. It accepts the original description of a lost person, as well as the description of the clue output by the Captioner (Stage 1). RAG is used to retrieve up to five associated pieces of guidance related to relevance, and these are integrated into the prompt.  For example, {\it``A clue is more likely to be relevant if it closely matches the known clothing or belongings of the lost person''.}  The LLM evaluates the lost-person and clue descriptions, using the guidance where helpful, to classify the clue’s relevance. It produces a categorical ranking (Very High, High, Medium, Low, None) together with a rationale as shown here:

\MyFrameBox{
\centering
\addtolength{\tabcolsep}{-3.8pt}
\begin{tabular}{@{}rL{6.6cm}@{}}
\textbf{Person:}  & \textcolor{InputColor}{An elderly man wearing glasses, an orange sweater, blue plaid pants, and gym shoes.}\\
\textbf{Clue:}  & \textcolor{InputColor}{A pair of broken glasses with a cracked lens on a rock.}\vspace{6pt} \\
\textbf{Relevance:}  & \textcolor{RoyalBlue}{High}\\
\textbf{Rationale:}  & \textcolor{RoyalBlue}{The broken glasses match the description of the lost man wearing glasses and were found beside a dirt path, suggesting possible alignment with his travel route.}
\end{tabular}
} 

\item {\bf Stage 3 - Task Planner:}  The task planner is responsible for planning the action to be taken based on the clue. 
Actions are defined as search tasks, where each task is focused on searching a specific named area (a subcluster $\kappa \in \mathcal{K}$). This includes boundaries of areas (e.g., shorelines of a river) or internal areas (e.g., a section of a forest or lake).  This stage takes three inputs describing the clue relevance, its rationale (output from Stage 2), and a description of terrain features in the vicinity of the clue. These include the terrain cluster in which the clue was found (\textit{on}), immediate neighbors (\textit{adjacent}), and all other clusters within a predefined radius (\textit{nearby}).  It also retrieves five pieces of task-planning advice from the RAG dataset, which are integrated into the prompt and used to inform the generated output action. As a result it outputs a prioritized list of possible actions, three of which are illustrated here: 

\MyFrameBox{
\centering
\addtolength{\tabcolsep}{-3.8pt}
\begin{tabular}{@{}rL{6.6cm}@{}}
\textbf{Relevance:}  & \textcolor{InputColor}{High}\\
\textbf{Rationale:}  & \textcolor{InputColor}{The broken glasses match the description of the lost man wearing glasses and were found beside a dirt path, suggesting possible alignment with his travel route.} \\
\textbf{Location:} & \textcolor{InputColor}{ON: Name of terrain feature where clue was found ADJACENT: List of immediately adjacent terrain features NEARBY: List of additional features within distance $D$} 
\vspace{6pt} \\
\textbf{Actions ($d$):}  & \textcolor{RoyalBlue}{Trail-10, Trail-11, Lake-5}
\end{tabular}
}

The task planner relies on the underlying scene model to interpret the context of the clue and to generate appropriate actions. 

\item{\bf Stage 4 - Triager:}
Finally, Stage 4 is responsible for determining how the plan will be enacted. Current options are limited to (a) permitting the sUAS that found the clue to perform the task, (b) sending the task to the drone pool for prioritization alongside other tasks to be performed, or (c) requesting operator review, with the possibility of overriding the decision. This stage accepts five inputs comprising the clue description, clue relevance, weather conditions, drone swarm status (i.e., location, health, and priority levels of all drones), LKP and Elapsed Time (ET). Here the LLM prompt is enriched with five pieces of triaging advice from the RAG dataset and outputs the triage decision.
%\begin{mdframed}[style=MyFrame]
\MyFrameBox{%
\centering
\addtolength{\tabcolsep}{-3.8pt}
\begin{tabular}{@{}rL{6.6cm}@{}}
\textbf{Clue:} & \textcolor{InputColor}{A pair of broken glasses with a cracked lens on a rock.}\\
\textbf{Relevance:}  & \textcolor{InputColor}{High}\\
\textbf{Weather:}  & \textcolor{InputColor}{Light=Bright, Weather=Snow, Temp=Hot}\\
\textbf{Drones: }  & \textcolor{InputColor} {RED:30mins/High; BLUE:10 mins/Med; AQUA:35mins/Med}\\
\textbf{Sighting: } & \textcolor{InputColor}{LKP: 64.328122, -20.516289, ET: 60 mins}\vspace{6pt} \\
\textbf{Assignment:}  & \textcolor{RoyalBlue}{Send task to Drone Pool}\\
\end{tabular}
} %\end{mdframed}

\item{\bf Post-Pipeline Enactment} When a task is assigned to a specific sUAS or to the sUAS pool, the assignee dynamically adapts from its current task to the new task. If the decision is made to engage humans, the sUAS continues its current action while a request for further analysis is made to a human operator, via an active Graphical User Interface.
\end{itemize}

While the CAP serves a critical purpose in interpreting clues and dynamically generating meaningful action plans, its use of foundational models means that it can potentially generate erroneous results, and therefore its decisions need to be checked by a runtime Cognition Envelope.  In our case example, the Cognition Envelope's external semantic validator ($M$) is comprised of the \psar and the heuristic-based MCE utility.

\section{Establishing a Cognition Envelope}
\label{sec:cognition_envelope}

Establishing a Cognition Envelope involves three key steps of requirements scoping, solution design, and validation.  In this section we focus on scoping and design.  Validation is deferred to Section \ref{sec:validation}, which describes how experimentation can be used within the software testing process to assess the efficacy of a Cognition Envelope.

\subsection{Scoping Envelope Responsibilities}
\begin{figure}
    \centering
    \includegraphics[width=\linewidth]{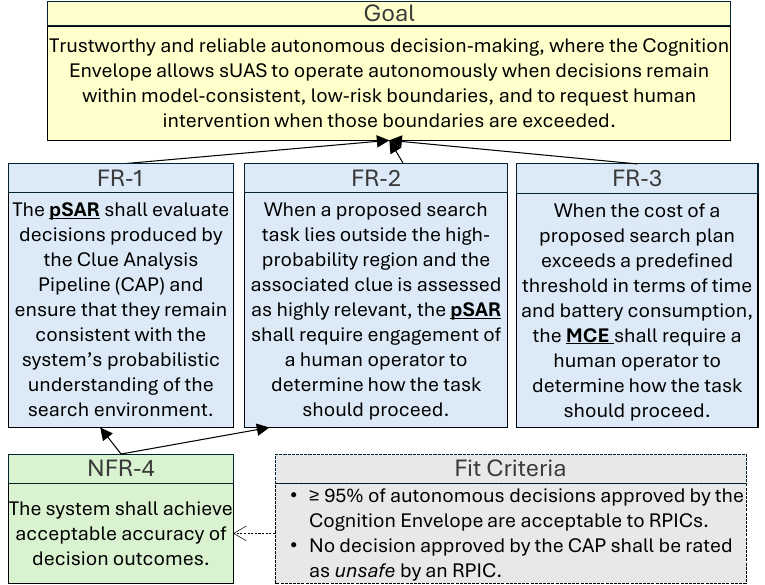}
    \vspace{-14pt}
    \caption{Decomposition of Cognition Envelope requirements specifying probabilistic coherence (pSAR) and the Mission Cost Evaluator (MCE).}
    \label{fig:hierarchy}
    \vspace{-14pt}
\end{figure}

We begin by defining the scope of the Cognition Envelope, thereby establishing clear guardrails around the CAP to support trustworthy and reliable autonomous decision-making. A  Cognition Envelope can be designed in two distinct ways. In a black-box configuration, it inspects only the final outcome of an LLM-guided process; while in a white-box configuration it looks inside the pipeline, examining the outputs of intermediate reasoning stages in order to detect flaws or inconsistencies before they influence the final output. In our example, the intermediate stages would include clue captioning, relevance assessment, and action selection.  In our worked example, we opted to treat the pipeline as a black box for two reasons. First, individual stages already incorporate localized meta-cognitive checks that support self-consistency, and second, developing cognition envelopes for each component would add significant design complexity and runtime cost. We capture our requirements and design decisions using a high-level quality-related goal, three functional requirements (FR-1, FR-2, FR-3), and one non-functional requirement (NFR-4), which is augmented with fit criteria \cite{robertson2012mastering}, as depicted in Figure~\ref{fig:hierarchy}. Responsibilities for satisfying the three requirements are distributed across the \psar, MCE, and human operators.  The \psar is charged with checking that a decision output by the CAP to search a specific search areas aligns with a reasonable probability of the lost person being in that search area, while the MCE checks for, and curbs decisions that are costly in terms of time and power consumption. Notably, the MCE is straightforward to implement and validate, while the \psar is more challenging. We therefore focus our attention on the \psar. 

\subsection{Designing \psar as a Semantic Model $M$} 
\label{subsec:semantic-model-m}
To design the \psar as the Cognition Envelope's semantic model $M$, we followed a systematic design process, exploring alternative architectural options and evaluating their ability to satisfy the stated requirements \cite{Nuseibeh2001TwinPeaks}. The final design uses a probability-based potential-field approach, which is well suited to modeling  established SAR best practices and for reasoning under uncertainty \cite{drones8120760}. Probability-based potential fields model spatial likelihood as a continuous surface in which higher values indicate regions with greater expected relevance to the search, allowing the system to reason about search priorities. As the full mathematical exposition of this model lies beyond the scope of this paper; we summarize its key principles and its role  in evaluating CAP’s candidate decision $d$ under evidence and mission context $e$ to produce the semantic acceptability signal $s := M(d,e)$.

\subsubsection{Probability of Area for Lost Person}
\psar builds on the concept of Probability of Area (POA) to represent the likelihood of the lost person being in different regions of the search area. Given mission context $e$, which consists of the last known point (LKP), elapsed time since establishing the LKP, a person profile, and environmental data, \psar produces a spatial belief (the POA field) over search subclusters $\kappa \in \mathcal{K}$. We then use this belief to evaluate CAP’s proposed search decision $d$, communicated as one or more recommended subclusters. The underlying distributions arise from two primary concepts of \emph{reachability} and \emph{affinity}.\vspace{2pt}

\noindent{$\bullet$~{\it Reachability Kernel.}}
Reachability estimates how physically accessible each terrain cell $c \in C$ is from the last known point (LKP). We compute an approximate minimum travel time to each cell over a weighted grid graph with 8-connectivity, where each edge cost is the distance between neighboring cell centroids divided by an effective movement speed. This effective speed is derived from terrain-dependent walking rates and adjusted for topographic and hydrographic effects, as well as physiological conditions (e.g. age, fitness/experience, weather exposure). The model is grounded in empirical studies documenting walking rates across diverse terrains and the effects of slope and water \cite{tobler1993three, koester2008lost}. The resulting minimum-time surface provides an interpretable map of likely movement corridors and natural barriers. To account for uncertainty in pace and elapsed time, we convert travel time into a smooth reachability weight $R(c)$. Cells reachable within the elapsed time are assigned high weight, while cells beyond the frontier decay continuously as the required travel time exceeds the elapsed time.

\noindent{$\bullet$~{\it Affinity Kernel:} }
Affinity captures how strongly a lost person’s movement is drawn toward or aligned with salient environmental features such as roads, trails, or shorelines. The {\it Affinity Field} $A(c)$ is implemented using smooth {\it radial basis functions (RBFs)} that assign high affinity to nearby features and gradually decrease influence with distance. \psar currently defines RBF fields for 11 types of features, including categories corresponding to terrain clusters $K_i$ and sub-clusters $\kappa$ for roads, waterways and their shorelines, woodlands and their boundaries, buildings, and open areas. Each feature is associated with an affinity strength parameter, and the total affinity for cell $c$ is the product of its feature-specific affinities. The resulting affinity surface $A(c)$ provides a probabilistic representation of environmental preference based on the lost-person profile \cite{jacobs2015terrain}. 

The probability of the lost person being in cell $c$ is represented by the probability-of-area field $p(c)$, computed as the product of physical reachability $R(c)$ and behavioral affinity $A(c)$.  Because CAP's outputs are recommended subclusters, we lift $p(c)$ to the search subcluster level by first aggregating probability mass within each subcluster, $p(\kappa)=\sum_{c\in\kappa} p(c)$ for $\kappa\in\mathcal{K}$, and then computing a \emph{size-normalized POA score}:
$\;q(\kappa)=\frac{p(\kappa)/|\kappa|}{\sum_{\kappa' \in \mathcal{K}} p(\kappa')/|\kappa'|}\;$,
where $|\kappa|$ denotes the number of unsearched cells in $\kappa$.

\subsubsection{Evidence Updates} After a clue has been detected with sufficient confidence, the POA is dynamically updated to reflect the new evidence. Each update adjusts spatial belief according to the type of information conveyed by the clue. For example, clues that confirm the victim’s past presence at a specific location refine probabilities locally around that point, while directional or feature-based clues reshape belief along implied movement paths or terrain structures. Finally, negative findings, such as area searches that are completed without finding a clue or the lost person, suppress probability in those areas. The adjustment process balances prior beliefs against the strength and reliability of the new evidence, ensuring that confident, well-supported clues meaningfully redirect the search while uncertain or low-value observations exert proportionally smaller influence. In this way, the model continuously integrates evolving field intelligence into a coherent, up-to-date spatial estimate of likely victim location.  Notably, when the CAP detects a clue and classifies it with medium relevance or higher, the POA evidence is updated, and the updated evidence is used in the Cognition Envelope to check the plans generated by the CAP.

\subsection{Evaluating CAP Plans with \psar}
We evaluate each CAP-generated search plan against the model’s probability-based ranking using two complementary signals: (a) the candidate’s \textbf{percentile rank} and (b) its \textbf{ratio to the top-ranked candidate}. Together with an uncertainty estimate $H_{\text{norm}}$ defined below, these quantities form the \textbf{semantic acceptability signal} $s$ produced by \psar for each proposed candidate $\kappa$, and they are passed to the Cognition Envelope gate $G$ (Definition~1).

The percentile rank indicates a search area's relative position within the sorted list (1.0 = top, 0.0 = bottom), while the ratio-to-top quantifies magnitude agreement as
\begin{equation}
\rho = \frac{q(\kappa_{\text{suggested}})}{q(\kappa_{\text{top}})},
\end{equation}
where $q(\cdot)$ is the normalized candidate score. These metrics convey orthogonal information: percentile captures relative ordering, whereas ratio-to-top detects cases where candidates are similarly ranked yet substantially weaker in absolute terms. Relying on percentile alone can be misleading in flat distributions, while the ratio-to-top ensures the proposed candidate is not orders of magnitude weaker than the best option.

To quantify model uncertainty, we compute normalized Shannon entropy over the candidate distribution:
\begin{equation}\label{eq:shannon_entropy}
H_{\text{norm}} = -\frac{1}{\log |\mathcal{K}|} \sum_{\kappa \in \mathcal{K}} q(\kappa) \log q(\kappa)
\end{equation}
where 0 indicates a sharply peaked distribution and 1 reflects complete uncertainty. This entropy score governs how strictly or loosely we evaluate CAP plans, balancing exploitation under confidence with exploration under uncertainty.

Each candidate $\kappa$ is assigned a gating outcome $G(\kappa)$ by the Cognition Envelope's gate $G$ (Definition~1) corresponding to one of three possible outcomes: \textbf{ACCEPT}, \textbf{ALERT}, or \textbf{REJECT}, determined by whether it satisfies entropy-adaptive thresholds on percentile rank $r$ and ratio-to-top $\rho$.
\begin{equation}\label{eq:accept_reject_criterias}
G(\kappa) =
\begin{cases}
\text{ACCEPT} & \text{if } r \geq r_{\text{accept}}(H) \text{ and } \rho \geq \rho_{\text{accept}}(H) \\
\text{ALERT} & \text{if } r \geq r_{\text{alert}}(H) \text{ or } \rho \geq \rho_{\text{alert}}(H) \\
\text{REJECT} & \text{otherwise}
\end{cases}
\end{equation}

The thresholds $r_{\text{accept}}(H)$, $\rho_{\text{accept}}(H)$, $r_{\text{alert}}(H)$, and $\rho_{\text{alert}}(H)$ adapt to entropy through linear interpolation between strict (low $H$) and loose (high $H$) regimes:
\begin{equation}
r_{\text{accept}}(H) = r_{\text{accept}}^{\text{strict}} + H \cdot (r_{\text{accept}}^{\text{loose}} - r_{\text{accept}}^{\text{strict}})
\end{equation}
and similarly for $\rho_{\text{accept}}(H)$, $r_{\text{alert}}(H)$, and $\rho_{\text{alert}}(H)$.

The \textbf{ALERT} band captures borderline cases that warrant human review: a candidate triggers an alert if it meets either the percentile or ratio threshold. When the model is confident, we apply stricter alert thresholds near $r_{\text{alert}}^{\text{strict}}$ and $\rho_{\text{alert}}^{\text{strict}}$, filtering out weak suggestions before they reach operators. As entropy rises and the probability field flattens, the thresholds shift toward $r_{\text{alert}}^{\text{loose}}$ and $\rho_{\text{alert}}^{\text{loose}}$, flagging more candidates for review and increasing human involvement under higher uncertainty.

%https://docs.google.com/spreadsheets/d/1TlLik9dJqKdcYyYLT-gI8lcQq6UZlv8jqjmC1xKDERI/edit?usp=sharing

\section{Validating Cognition Envelopes}
\label{sec:validation}
Establishing and validating the effectiveness of a Cognition Envelope, especially one that includes probabilistic safeguards such as \psar, is non-trivial because performance depends not only on algorithmic correctness but also on how well the model captures operational constraints, mission dynamics, and less tangible concepts such as human expectations. Traditional unit tests are insufficient for such systems, as correctness cannot be assessed solely through input–output verification. Instead, validation must rely on large-scale simulations that capture environmental variability, operational constraints, and the interaction between autonomous reasoning and human oversight.
We therefore set up the validation process guided by the following research questions.
\\[2pt]\noindent
\textbf{RQ1:} Which stages of CAP’s decision-making process require \emph{external} cognition checks as part of the Cognition Envelope, beyond CAP’s internal meta-reasoning?
\\[2pt]\noindent
\textbf{RQ2: } How effectively do Cognition Envelope safeguards enforce autonomy constraints when evaluating CAP-generated decisions?

\subsection{Experimental Design for Validating \psar}
\begin{figure}
    \centering
    \includegraphics[width=\linewidth]{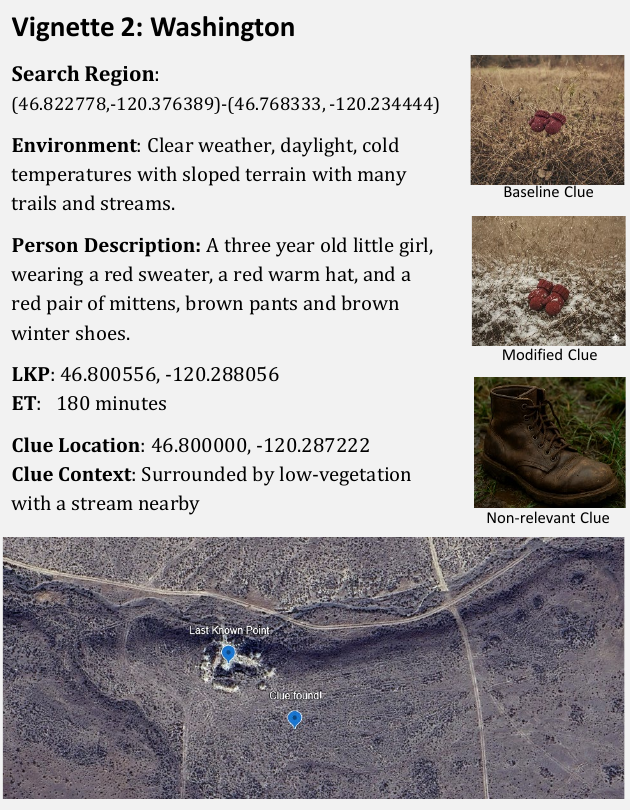}
    \caption{A Vignette describes the SAR mission, establishing the context for each CAP decision. It hosts a test suite, varied by clues, environmental conditions, and mission status.}
    \label{fig:vignette}
   \vspace{-18pt}
\end{figure}

To address these research questions we designed our evaluation around a set of 10 'vignettes', each representing a unique SAR mission. An example, describing the search for a 3 year old girl lost in Kittatas, Washington is shown in Figure \ref{fig:vignette}. The vignettes were developed independently by three team members not otherwise involved in developing the CAP or its underlying \psar model \cite{972722}. 

\subsubsection{Test Design} First, the team identified five historical accounts of prior SAR missions, each in a unique region.  They then constructed two \emph{vignettes} per region resulting in a total of ten vignettes (V1,V2,...,V10). The first vignette for each region was anchored in the documented report of the SAR event; however, because publicly available details were often sparse, they augmented the scenarios with plausible but fictional details (e.g., clothing and weather conditions) to support realistic clue interpretation and decision-making. They also carefully inspected the available terrain data and any known coordinates associated with each historical account, and used this information to place the clue at a meaningful location. The second vignette for each region was constructed around a hypothetical scenario. Ultimately, each vignette included a profile of the lost-person describing their age, gender, and clothing, and the elapsed time (ET) since the previous sighting at a last known point (LKP). The region description included search boundaries, terrain, weather conditions, and lighting. The sUAS swarm data described the number, current location, and health (e.g., battery level) of available aircraft. 

Each vignette included a test suite composed of seven distinct tests, labeled $T0$--$T6$. Test $T0$ established the baseline by defining a clue (e.g., lost mittens) relevant to the lost person. All remaining tests were variants of this baseline. $T1$ used a distorted version of the original clue (e.g., blurred, obfuscated, or damaged), while $T2$ introduced a non-relevant clue (e.g., a broken bicycle when searching for a lost child). $T3$  represented modifications to sUAS-related or mission parameters, while $T4$ included environmental and weather changes from the baseline scenario (often impacting the appearance of the clue). For these five initial tests, the clues were placed at locations within expected range of the search given the LKP and elapsed time, meaning that the \psar was likely to compute a non zero POA value for the terrain features around the clue's location. In contrast,  for tests $T5$ and $T6$ the clues were placed at remote locations from the LKP, where they had a higher likelihood of being out of the predicted search area. However, while real-world clues are varied in nature, including examples such as footprints, crushed grass, cell-phone beeps, lost clothing or dropped water bottles, all of our tests focused on objects such as clothing or other personal items that are within the scope of current CV models and VLM. We leave other forms of clue detection, such as footprints and crushed grass, to future work. 

In addition to these variants, each test was executed across a range of  {\it elapsed time} (ET) values of ET = 10, 20, 40, 60, and 90 minutes. This impacted the boundaries of the POA frontiers, which tend to expand at higher ET values, increasing POA around the clue location over time. Conversely, at low ET values, some clues are likely to yield negligible POA. This resulted in 350 unique tests, composed of 10 vignettes (2 for each of 5 regions), 7 tests per vignette and 5 ET variants per test.

\subsubsection{Example Vignettes and Tests by Region} We provide summarized examples of vignettes for each of the five regions, with the complete set of test cases provided online (see \url{https://github.com/SAREC-Lab/CAIN2026-CognitionEnvelopes}). 

\begin{itemize}[leftmargin=*]
\setlength{\itemsep}{2pt}
    \item \textbf{Rock River, Illinois} [V1.\textsc{Test-0}] %[Test 1, V0]
    
    \textit{Search region:}  
    (41.470261, -90.531366) to (41.446453, -90.403751).  
   \textit{ Lost person:} A teenage boy wearing a red shirt, red bucket hat, brown shorts, sneakers, and glasses.  
    \textit{Clue found:} A red hat floating in the water.
   \textit{ News report:} \url{https://tinyurl.com/sar-rockriver}
   
    \item \textbf{Kittitas, Washington} [V2.\textsc{Test-0}] %Test 2, V0]
    
    \textit{ Search region:} (46.822778, -120.376389) to (46.768333, -120.234444).
    \textit{Lost person:} A three year old little girl, wearing a red sweater, a red warm hat, and a red pair of mittens, brown pants and brown winter shoes.  
    \textit{Clue found:} A pair of red mittens on the ground.
    \textit{News report:} \url{https://tinyurl.com/sar-washington}

    \item \textbf{Mesa County, Colorado} [V3.\textsc{Test-0}]%Test 3, V0]\\
    
   \textit{Search region:} (38.962223, -108.333285) to (38.956922, -108.314291).  
   \textit{ Lost person:} A moderately experienced hiker wearing a purple jacket, purple hat with orange goggles, black pants, hiking boots, a large black backpack, and carrying two hiking sticks.  \textit{Clue found:} Smoke rising above the trees. 
   \textit{ News report:} \url{https://tinyurl.com/sar-mesa}

    \item \textbf{Pulaski, Arkansas} [V4.\textsc{Test-0}]%Test 4, V0]
    
    \textit{Search region:} (34.590254, -92.266513) to (34.552362, -92.189868).
    \textit{Lost person:} Elijah, a 16-year-old male wearing a woolly orange-brown hat, orange sweater, dark blue–yellow striped pants, and gym shoes. Last seen arriving at a church on a purple bicycle. 
    \textit{Clue found:} A brownish knit hat on the gravel. \url{https://tinyurl.com/sar-arkansas}

    \item \textbf{Los Angeles, California} [V5.\textsc{Test-0}]%[Test 5, V0]

    \textit{Search region:} (34.265727, -118.112756) to (34.217358, -118.048056). 
   \textit{ Lost person:} An elderly, experienced hiker and camper wearing a purple jacket, purple winter hat with black and orange goggles, black pants, hiking boots, a large black backpack, and carrying two hiking sticks.  
   \textit{Clue found:} A pair of goggles on sandy colored rock.
    \textit{News report:} \url{https://tinyurl.com/sar-california} 
\end{itemize}

\subsubsection{Clue Surroundings} Additionally for experimental purposes, for each unique clue location we retrieved the set of named search areas surrounding the clue from the terrain model. These included (1) the terrain subcluster on which the clue was placed, (2) the immediately adjacent subclusters, and (3)  additional subclusters within a predefined radius of 10 cells.  An example showing the named search areas for the baseline clue in Figure \ref{fig:vignette} is shown below, indicating that a large vegetation region (Low\_Vegetation-42) was decomposed into smaller subsections, and that a stream was also nearby.

\MyFrameBox{%
  \centering
  \begingroup
  \setlength{\tabcolsep}{2pt}
  \begin{tabular}{@{}r l@{}}
    \textbf{On:} &
      \parbox[t]{6.6cm}{\raggedright Low\_Vegetation-42-edge-82} \\[4pt]
    \textbf{Adjacent to:} &
      \parbox[t]{6.6cm}{\raggedright
        Low\_Vegetation-42-edge-26, Low\_Vegetation-42-edge-71, Low\_Vegetation-42-edge-7,
        Low\_Vegetation-42-edge-28, Low\_Vegetation-42-edge-145, Low\_Vegetation-42-edge-163} \\[4pt]
    \textbf{Nearby:} &
      \parbox[t]{6.6cm}{\raggedright
        Low\_Vegetation-42-edge-80, Low\_Vegetation-42-edge-82, Stream\_River-7-edge-3}
  \end{tabular}
  \endgroup
}

Notably for future real-world deployments, this data would be retrieved dynamically by sUAS at the time that a clue is discovered.

\begin{figure}
    \centering
    \includegraphics[width=\linewidth, trim=0 .4cm 0 0, clip]{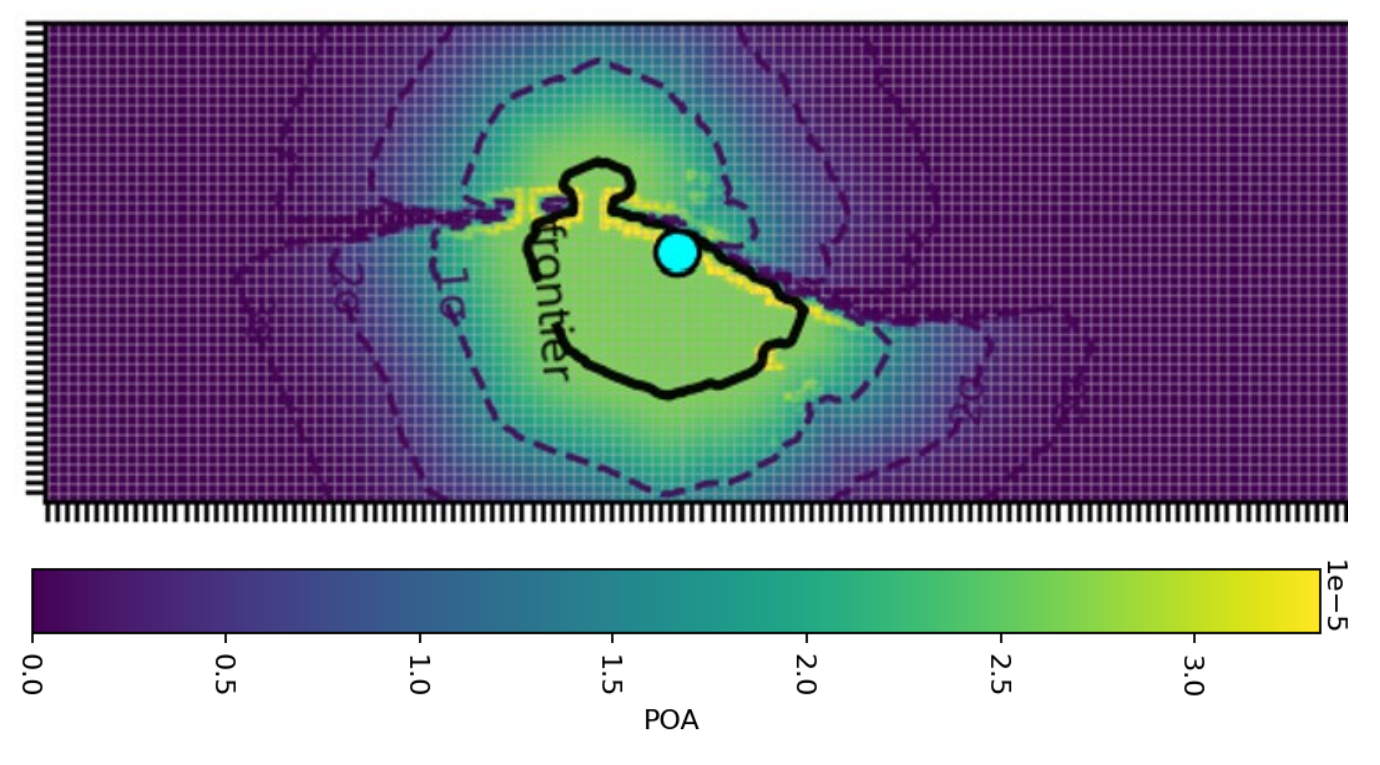}
    \caption{Visualization of the Probability of Area (POA) from the LKP (aqua circle). Probabilities are computed based on reachability and affinity and influenced by time elapsed since the Last Known Point. This example shows how a river impedes reachability; however,  other types of impedance such as cliffs or marshland can have similar effects.}
    \vspace{-12pt}
    \label{fig:poa}
\end{figure}

\subsection{Experimental Infrastructure}
We implemented the Clue Analysis Pipeline using GPT-4.0 (the latest model available as a python API at the time) and accessed through the OpenAI API, with inference performed at a temperature of 0.2 to ensure consistent outputs. The system was built with the LangChain framework (langchain, langchain-community, langchain-openai) and used a FAISS vector store (faiss-cpu) for retrieval-augmented grounding. \psar was implemented in Python via numpy and standard libraries. 

To execute each of the 350 tests, we first ran the CAP using clues and other data defined in each test case. CAP outputs were recorded in a JSON file. This part of the experiment was conducted on a system with an Intel Core Ultra 7 256V (2.20 GHz) processor, 16 GB RAM, and a 64-bit Windows operating system, with an average runtime of 12 seconds per clue. 
The \psar part of the experiment was run on a server with 26 Intel Xeon processors, 159GB RAM, and an Ubuntu 22.04 operating system. The execution time of \psar is highly dependent on the overall size of the region, ranging from an average runtime of 3 seconds for Region 3 ($\sim$ 1 $km^2$) to an average runtime of 1 minute for Region 2 ($\sim$ 66 $km^2$).  The input to pSAR was provided as json files containing the general scenario information and terrain data $e$ as well as the output $d$ from CAP. pSAR initially processes the terrain data to build a scene (discussed in Sec. \ref{sec:clues}), and incorporate the priors of the scenario to build the initial POA map (Sec. \ref{subsec:semantic-model-m}).  When given a valid clue location, pSAR updates its probabilities to incorporate this information.  Finally, the results are saved in an output json file containing the gating outcomes $G$ for all CAP candidates and clues.

\begin{figure}
    \centering
    \includegraphics[width=\linewidth]{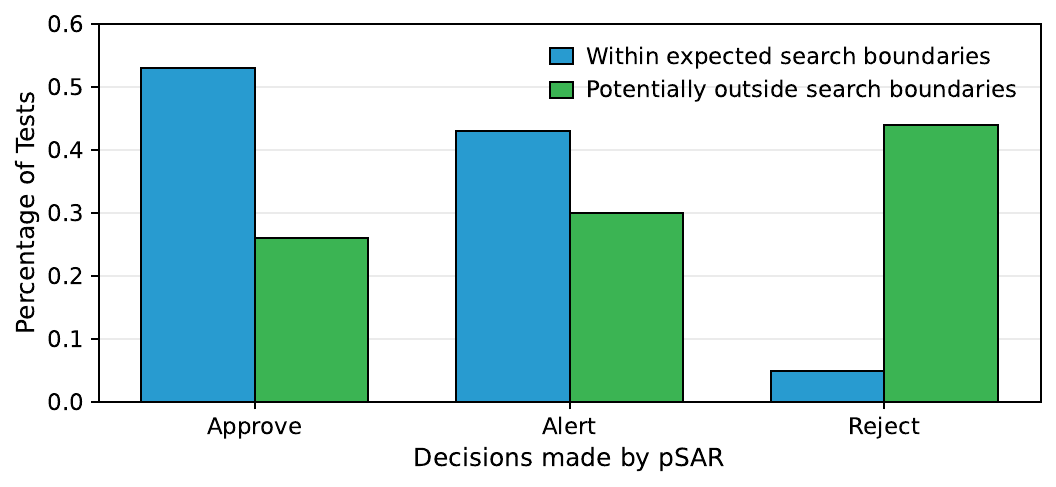}
    \caption{\psar Decisions when using POA scores without updating based on the clue. In this case the majority of decisions within the current search area are approved, while many outside the area are rejected despite the presence of the clue.}
    \label{fig:results1}
    \includegraphics[width=\linewidth]{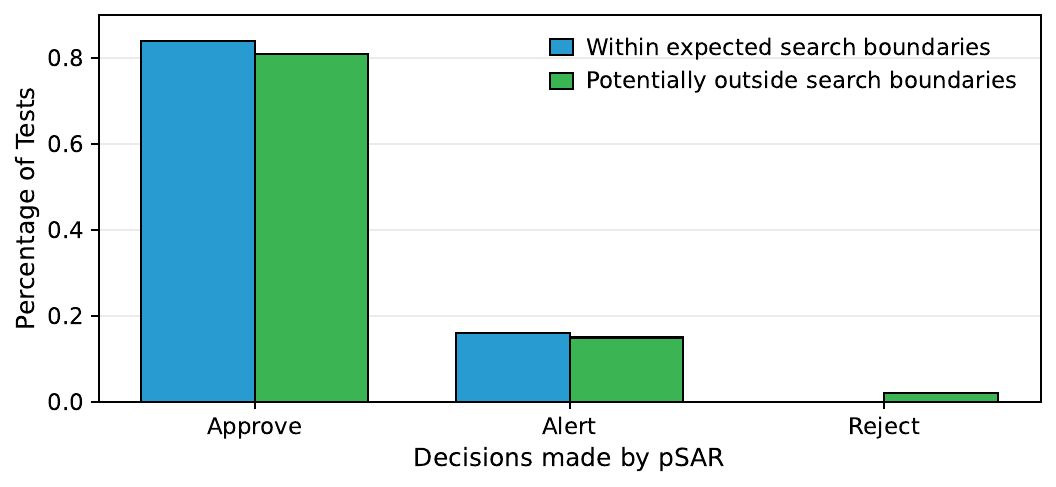}
    \caption{\psar decisions made after the POA was updated, potentially supporting increased decision autonomy. }
    \label{fig:results2}
\end{figure}
\subsection{Results and Analysis} 
After executing all 350 tests we analyzed results in order to address our two research questions.

\subsubsection{RQ1: Which stages of CAP’s decision-making process require external cognition checks as part of the Cognition Envelope, beyond
CAP’s internal meta-reasoning?}
To address {\bf RQ1} we evaluated CAP outcomes to determine whether the CAP appropriately categorized the clues as {\it relevant} or {\it non-relevant}. Each test suite included one variant ($V2$) that represented a non-relevant clue. For example in Figure \ref{fig:vignette}, the vignette included an old boot which was clearly not relevant to the lost child. Results from the experiment showed that in every case the non-relevant clue was successfully rejected. Further, in eight out of the ten test suites, all relevant clues were marked as relevant. Upon inspection, in one case the CAP failed to mark a clue as relevant due to severe occlusion (caused by a bird flying in-front of the clue in the image), and in the secondary case the CAP failed to understand the significance of campfire smoke. However, this problem could be remedied in future implementations through adding additional guidance to the RAG dataset. Overall Stages 1 and 2 of the CAP pipeline achieved 95\% accuracy with 47 true positives, 0 false positives, 10 true negatives, and 3 false negatives. All detected clues rejected by CAP are surfaced to the operator for optional review.

These results indicate that CAP’s internal meta-reasoning is largely sufficient for supervising clue interpretation and relevance assessment, and therefore external Cognition Envelope checks provide limited additional value at these early stages. In contrast, even when a clue is correctly interpreted and deemed relevant, subsequent action-selection decisions (Stages~3--4) must determine whether the proposed search task is plausible given environmental constraints, elapsed time since the LKP, and the mission cost of redirecting limited sUAS resources. These judgments depend on probabilistic reasoning about mobility and reachability, and on policy-driven autonomy constraints (e.g., preventing expensive or low-plausibility searches from being enacted without oversight). As shown in Figure~\ref{fig:pipeline}, the Cognition Envelope is therefore most needed at the task planning and triggering stages, where \psar-derived plausibility assessments and cost-aware autonomy checks determine whether CAP’s proposed action can be safely approved for execution or must be escalated to the human-on-the-loop.

\subsubsection{RQ2: How effectively do Cognition Envelope safeguards enforce autonomy constraints when evaluating CAP-generated search actions? }

The remainder of our evaluation focuses on the cases where clues were marked as relevant and search plans were generated and forwarded to the \psar for evaluation.  To perform this evaluation we classified test cases into two categories. Group 1 included those for which the clues were placed at vicinities likely to be within the currently active search region (i.e., tests $T0$-$T4$ for each Vignette), while Group 2 included tests where clues were placed at more extreme coordinates and were more likely to be outside the active search region (Tests $T5$,$T6$ for each Vignette). We evaluated the \psar performance under two distinct scenarios, first when search plans were evaluated using existing POA values without updating scores based on the location and potential relevance of the found clue, and second, after pSAR updated its probabilities to reflect the discovered clue. We ran a total of 360 unique tests, comprised of 240 from Group 1 and 120 from Group 2.

Figure \ref{fig:results1} shows results obtained when the \psar did not update the underlying probability model based on the clue.  Here, for  Group 1, 53\% of plans were approved, 43\% were directed to human operators for review, and only 5\% were rejected. This contrasted with Group 2, where only 26\% of plans were approved, 30\% resulted in alerts, and 44\% were rejected. In contrast, Figure \ref{fig:results2} reports results when \psar updated its underlying models to reflect the discovered clue prior to analyzing the planned search tasks. This had a strong effect of increasing probabilities around the area of the clue, thereby increasing the approval rate of search plans in the vicinity of the clue for all test cases, whether in Group 1 or Group 2.  In both cases
a manual inspection of outcomes showed that the three major inhibitors were distance from the LKP, Elapsed Time (which varied under the different treatments), and terrain barriers such as rivers or cliffs, all of which impacted current POA computations. This highlights the 
importance of updating the \psar model to incorporate the discovered clue’s location, which increases the computed probability of the person being in nearby areas and in regions naturally reachable from that point. Updating the model correspondingly increases approval rates for clue-related plans generated by the CAP, and may enable higher levels of autonomy. When approvals are high, the MCE component remains essential by ensuring time and power costs stay within thresholds, especially for distant clues where travel costs are substantial.

Overall, these results  provide a proof-of-concept that Cognition Envelopes can deliver meaningful checks on foundation-model decisions in an operational sUAS setting.

\section{Open Challenges for Cognition Envelopes}
\label{sec:roadmap}
This paper has presented the notion of Cognition Envelopes as a means for independently checking whether LLM/VLM reasoners produce valid plans. We have further demonstrated the effectiveness of a Cognition Envelope based on a combination of a probabilistic approach reflecting the likelihood of a person having been in the vicinity of the clue, and a complementary heuristic approach for measuring resource consumption. In this section we discuss  generalizability of the approach and open challenges for Software Engineering of Cognition Envelopes.

\subsection{Generalizability of Cognition Envelopes}
Cognition Envelopes provide an independent check on the soundness of LLM-generated decisions, and can be realized through techniques such as probabilistic validation, rule-based constraints, knowledge-consistency checks, ROI thresholding, and human engagement triggers. While a full generalizability study is beyond this paper’s scope, we now illustrate how Cognition Envelopes could be applied in three additional systems.

Our first example considers UAV exploration in partially observed environments. Xie et al. \cite{xie2024} use an LLM to reason about unseen regions and propose navigation paths. In terms of the $\langle d, e, M, s, G \rangle$ tuple in Definition~1, the candidate decision $d$ is the proposed next exploration path or waypoint sequence, and the evidence $e$ includes the current belief or partial map, recent observations, and mission constraints. The semantic model $M$ evaluates $d$ under $e$ to produce a semantic acceptability signal $s:=M(d,e)$. Here, $s$ can be instantiated as a pair $(v,c)$, where $v$ is a scalar expected information gain estimate for the proposed path computed from the belief state and $c$ is a confidence measure derived from map coverage or uncertainty. The gate $G$ then accepts proposals when $v$ exceeds a threshold and $c$ is sufficiently high; otherwise, it rejects or escalates.

Our second example considers multi-UAV formation control guided by multi-modal LLM outputs. Liu et al.~\cite{Liu_2024} use a pre-trained LLM to map user intent and perceptual context to target positions for a UAV team. Here, the candidate decision $d$ is the proposed formation target assignment and the evidence $e$ includes the intended formation together with the current swarm state (e.g., per-UAV poses and separations). A Cognition Envelope can instantiate $s$ using a non-LLM geometric validator $M$ that (i) measures deviation from the desired formation geometry, and (ii) checks feasibility constraints (e.g., spacing and connectivity) via a proximity graph over the proposed targets. The gate $G$ accepts when both the formation error and constraint violations fall below thresholds, and otherwise rejects or escalates.

Our final example moves beyond UAVs to precision oncology, where Benary et al.~\cite{benary2023} evaluate conversational LLMs on genomics-driven cancer vignettes and the therapies and trials they propose. Here, the candidate decision $d$ is the set of proposed treatment options for a tumor--alteration context, and the evidence $e$ includes the cancer type, molecular findings, and relevant clinical constraints. A Cognition Envelope can instantiate the semantic acceptability signal $s$ using a non-LLM precision oncology evidence engine $M$ grounded in molecular tumor board workflows, which maps each proposed option to structured variant-drug knowledge bases and guideline corpora to quantify actionability. Concretely, $s$ can be instantiated as a pair $(\ell,\gamma)$, where $\ell$ is an evidence tier assigned to the proposed option in the given tumor context and $\gamma$ is a guideline concordance score indicating whether the option is standard-of-care or investigative. 

Taken together, these worked examples suggest that Cognition Envelopes are potentially applicable across a broad-range of LLM-supported CPS.  Clearly further work is needed to deploy and validate their use. 

\subsection{Open Challenges}
Our experience engineering a Cognition Envelope for sUAS highlighted  open Software Engineering challenges as discussed below.
\begin{itemize}[leftmargin=*]
\item{\bf Scoping the Cognition Envelope}:
A cognition envelope may combine multiple interacting validators that work together to establish boundaries around a decision-making process. Clear scoping of Cognition Envelope's role Designing a trustworthy Cognition Envelope requires clear scoping; otherwise it risks providing false assurance instead of reasoning-level protection.~~{\it Mitigation:} Specify inputs, checks, decision roles, and outputs, including what the envelope verifies, vetoes, or only monitors.
\item {\bf Ground-Truth Alignment under Uncertainty}:
Cognition Envelopes depend on evidence from sensors, logs, or runtime models, but that data is often incomplete or noisy, and may overlap with the LLM’s evidence, introducing confirmation bias. 
{\it Mitigation:} Use meta-monitor to assess reliability of the Cognition Envelope at runtime through quality analysis of underlying evidence sources. When self-confidence is low, trigger conservative modes or human oversight.
\item{\bf Verifying the Verifier:}
Internal probabilistic reasoning, symbolic constraints, or logic within the Cognition Envelope can be flawed, silently allowing unsafe actions or blocking valid ones. 
{\it Mitigation:} Treat the envelope as a first-class component: test inference paths, simulate edge cases, and audit approval/rejection behavior.
\item{\bf Human Engagement:}
Hand-off criteria and interfaces must avoid both over-interruption and late intervention. 
{\it Mitigation:} Use measurable thresholds (confidence/uncertainty/anomalies) and provide brief, actionable rationales when decisions are escalated to human overseers. 
\item{\bf Explainability and Auditability:}
Approvals, rejections, and hand-offs must be inspectable by engineers and regulators. 
{\it Mitigation:} Log structured decision records (inputs, applied checks, confidence, outcome).
\item{\bf Meta-Cognition vs.\ Cognition Envelopes:}
A key open question is when to use meta-cognition, versus cognition envelopes, or hybrid combinations of both. 
{\it Mitigation:} Develop patterns that define the contexts under which each should be used, and explore opportunities for hybrid applications.

\end{itemize}

These challenges establish a foundation for future work on envelope specification, assurance, adaptation, and long-term validation in LLM-enabled CPS.
\section{Related Work}
\label{sec:related}

Recent work increasingly integrates LLMs and VLMs into autonomous UAV operations~\cite{hu2025llvm,llm4uavsurvey,tian2025uavs}, enabling capabilities such as vision-language waypoint generation for search and rescue~\cite{uavvlrr}, language-driven navigation for object search~\cite{liu2023aerialvln,flysearch}, and edge-based in-context learning~\cite{prompts2protection}. While these systems demonstrate adaptive planning from complex operational context, they largely lack systematic reasoning safeguards.

LLM hallucinations remain a core limitation, driven in part by training incentives that favor confident outputs over calibrated uncertainty \cite{openai2025hallucinate}. In safety-critical autonomy, such errors can yield incorrect situational assessments and unsafe decisions \cite{wang2024hallucination,sood2025paradigm}, motivating the need for independent reasoning safeguards at runtime.
Runtime assurance provides one promising foundation for such safeguards in learning-enabled systems that cannot be fully verified offline \cite{schierman2020runtime,hobbs2023runtime}. Related efforts translate natural-language intent into formally checkable constraints for runtime enforcement \cite{yang2024plug,liu2024lang2ltl}. However, these approaches primarily target operational safety, and do not directly validate the semantic correctness or evidential grounding of LLM-generated decisions—precisely the gap addressed by Cognition Envelopes.

Probabilistic reasoning and Bayesian inference are long-standing tools for managing uncertainty in robotics, and have been applied to UAV decision-making under uncertainty \cite{di2018threat,hossain2022assessing,kohaut2023mission}. Their ability to represent uncertainty and risk provides a natural basis for assessing the evidential soundness of LLM-generated plans.

Metacognition has also been explored to improve LLM reliability via self-reflection and self-critique \cite{bilal2025meta,conway2024toward,renze2024self}. However, studies show that models often remain overconfident, fail to recognize knowledge gaps, and exhibit fragile internal reasoning \cite{deepmind2023large,griot2025large}, suggesting that internal self-critique alone is insufficient for high-assurance settings and motivating external validation.

Related work on LLM guardrails constrains model behavior through input/output filters, embedding-based classifiers, or LLM-as-a-judge mechanisms \cite{nvidia2023nemo,dong2024building}, but typically does not validate whether decisions are evidence-grounded in the operational context.
\section{Threats to Validity}
\label{sec:threats}
The work in this paper has presented the concept of a Cognition Envelope. However, there are several threats to validity.

\noindent{\bf Construct Validity}. While we have discussed potential generalizability across CPS applications, this paper focused on a single Cognition Envelope applied to a single LLM-based pipeline. Moreover, the Cognition Envelope we developed and validated included only two techniques based on probabilistic reasoning and rule-guided human engagement. Further work is needed to explore additional envelope types and domains.

\noindent{\bf Internal Validity}. Our primary experiments used decision-related vignettes rather than actual missions, since realistic missions in our high-fidelity simulation can take from 15 minutes to over an hour. Because our focus was decision checking, vignettes enabled a larger number of scenarios with variants and injected mutants, and therefore more data for empirical analysis.

\noindent{\bf External Validity}. A threat to external validity stems from the domain specificity and limited evaluation environment of our study. The Cognition Envelope and its underlying \psar\ model were developed and tested primarily in simulation, and generalizability to broader CPS domains or field-deployed systems has not yet been verified. In future work, we plan to deploy the complete system both in simulation and on physical sUAS platforms already equipped for onboard reasoning and decision checking.

\noindent{\bf Conclusion Validity}. Our experiments did not include a head-to-head comparison against a pipeline-level meta-cognition layer. Our hypothesis, not yet evaluated, is that meta-cognition would suffer from the very problems we aim to address, and therefore an external and orthogonal approach such as \psar is essential. Future experiments are needed to compare these approaches across realistic mission contexts and reasoning tasks.

\section{Conclusions}
\label{sec:conclusions}
This paper has proposed Cognition Envelopes as an independent mechanism for checking decisions produced by systems that employ foundational models such as LLMs and VLMs for reasoning, inference, and planning. Rather than guaranteeing complete correctness, Cognition Envelopes evaluate each decision against well-scoped criteria that define acceptable reasoning boundaries. Within these boundaries, they can provide specific assurances; for example, in our \psar model, decisions falling outside probabilistic expectations are automatically redirected for human review. Our experimental results demonstrate that this approach can detect flawed or unsafe decisions while maintaining operational continuity, providing a practical step toward more trustworthy, transparent, and accountable AI-enabled systems. Finally, the paper has outlined open Software Engineering challenges that form an initial roadmap for advancing research and practice in the development, validation, and assurance of Cognition Envelopes.\vspace{12pt}

\noindent {\bf Supplemental materials:}\newline \url{https://github.com/SAREC-Lab/CAIN2026-CognitionEnvelopes}.

% Old link: https://tinyurl.com/cain2026-cog-env

\begin{acks}
Funding for foundational aspects of this project was provided by the USA National Science Foundation under grant \# 1931962.
\end{acks}

\bibliographystyle{ACM-Reference-Format}
\bibliography{_bibby}
\end{document}